%% file: main.tex
\documentclass[runningheads]{llncs}
\usepackage{eccv}
\usepackage{eccvabbrv}
\usepackage{graphicx}
\usepackage{booktabs}
\usepackage[accsupp]{axessibility}  
\usepackage{hyperref}
\usepackage{amsmath}
\usepackage{amssymb}
\usepackage{color}
\usepackage{multirow}
\usepackage[dvipsnames]{xcolor}
\usepackage{colortbl}
\usepackage{arydshln}
\usepackage{tabularx} 
\usepackage{wrapfig}
\usepackage{tikz}
\usepackage{comment}
\usepackage[bottom]{footmisc}
\usepackage{orcidlink}

\begin{document}

\title{ReSyncer: Rewiring Style-based Generator for Unified Audio-Visually Synced Facial Performer}

\titlerunning{ReSyncer}

\author{
Jiazhi Guan
\inst{1,2}$^*$ \and
Zhiliang Xu\inst{2}$^*$ \and
Hang Zhou\inst{2}$^{\dag}$ \and
Kaisiyuan Wang\inst{2} \and
Shengyi He\inst{2}\and
Zhanwang Zhang\inst{2} \and
Borong Liang\inst{2} \and
Haocheng Feng\inst{2} \and
Errui Ding\inst{2} \and
Jingtuo Liu\inst{2}\and
Jingdong Wang\inst{2} \and
Youjian Zhao\inst{1,3}$^{\dag}$ \and
Ziwei Liu\inst{4}
}

\authorrunning{J. Guan et al.}

\institute{
$^1$BNRist, DCST, Tsinghua University \quad
$^2$Baidu Inc. \quad \\
$^3$Zhongguancun Laboratory \quad
$^4$S-Lab, Nanyang Technological University \\
\email{guanjz20@mails.tsinghua.edu.cn, zhouhang09@baidu.com}
}

\maketitle
\def\thefootnote{*}\footnotetext{Equal Contribution.}
\def\thefootnote{\dag}\footnotetext{Corresponding Authors.}
\def\thefootnote{\arabic{footnote}}

\begin{center}
 \centering
 \includegraphics[width=\textwidth]{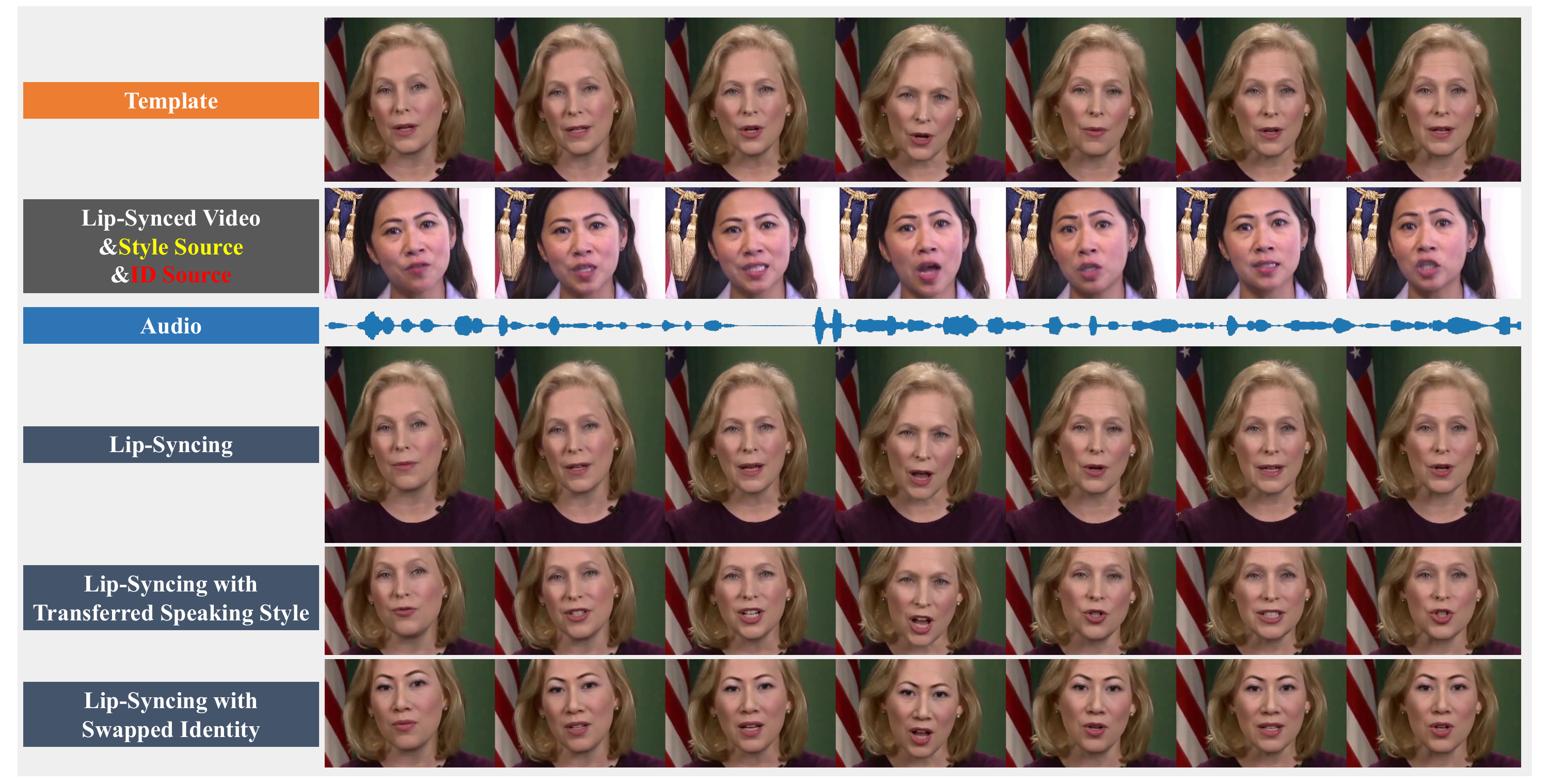}
\captionof{figure}{
\textbf{Lip-Synced/Speaking-Style-Transferred/Face-Swapping Results by ReSyncer.} 
Our method not only produces high-fidelity lip-synced video according to audio but can further transfer the speaking style and identity of any target person.
}
\label{fig:teaser}
\end{center}

\input{sec/0_abstract}    
\input{sec/1_intro}
\input{sec/2_related}
\input{sec/3_method}

\input{sec/4_exp}
\input{sec/5_conclusion}

\bibliographystyle{splncs04}
\bibliography{main}

\clearpage
\input{sec/99_appendix}

\end{document}

%% file: sec/0_abstract.tex
\begin{abstract}

Lip-syncing videos with given audio is the foundation for various applications including the creation of virtual presenters or performers. While recent studies explore high-fidelity lip-sync with different techniques, their task-orientated models either require long-term videos for clip-specific training or retain visible artifacts. In this paper, we propose a unified and effective framework \textbf{ReSyncer}, that synchronizes generalized audio-visual facial information. The key design is revisiting and rewiring the Style-based generator to efficiently adopt 3D facial dynamics predicted by a principled style-injected Transformer. By simply re-configuring the information insertion mechanisms within the noise and style space, our framework fuses motion and appearance with unified training. Extensive experiments demonstrate that ReSyncer not only produces high-fidelity lip-synced videos according to audio, but also supports multiple appealing properties that are suitable for creating virtual presenters and performers, including fast personalized fine-tuning, video-driven lip-syncing, the transfer of speaking styles, and even face swapping.
Resources can be found at \url{https://guanjz20.github.io/projects/ReSyncer}.
\end{abstract}

%% file: sec/1_intro.tex
\section{Introduction}
\label{sec:intro}

Generating lip-synced facial videos with speech audio is a foundational technique for various applications including the dubbing of movies and the creation of virtual presenters or performers in the form of realistic humans. 
The needs in virtual performers' creation can hardly be comprehensively met. 

While great attention has been paid to creating dynamic lip-synced videos from a few photos~\cite{jamaludin2019you,chen2019hierarchical,zhou2019talking,zhou2021pose,zhou2020makelttalk,wang2021audio2head,wang2022one,zhang2023sadtalker}, previous results are still far from realistic.
To generate life-like videos, researchers have made explorations~\cite{prajwal2020lip} with GANs~\cite{thies2020neural,cheng2022videoretalking}, NeRF~\cite{guo2021adnerf,liu2022semantic,tang2022real,shen2022dfrf}, and Diffusion models~\cite{shen2023difftalk} to edit existing videos. Despite their efforts, most methods rely on a relatively long-term clip of video for training, making them less flexible and efficient. A few methods~\cite{prajwal2020lip,sun2022masked,guan2023stylesync,cheng2022videoretalking} enable generalized local editing without training. However, they still tend to create visible flaws on high-quality videos~\cite{prajwal2020lip,sun2022masked}
Particularly, StyleSync~\cite{guan2023stylesync} supports not only arbitrary video modeling but can adapt to the speaking style of a target person with short-clip fine-tuning, which balances the generation quality and efficiency.
Nevertheless, their results can still be easily affected by mouth movements on their conditional reference frames, leading to unstable results.

The previous cases reveal the limitations of using low-dimensional audio information to directly modify high-dimensional visual data.
On the other hand, intermediate representations like 3D facial models are widely used, but their potential as spatial guidance for efficient, generalized editing remains under-explored. Using only 3DMM~\cite{blanz1999morphable} coefficients~\cite{thies2020neural,Deng_2020_CVPR,Gafni_2021_CVPR,Deng_2022_CVPR,ma2023styletalk,li2021write} also suffers from the cross-domain information injection problem.
While textured rendering offers a promising alternative~\cite{wu2021imitating,thies2020neural,chen2020talking}, its implementation complexity and susceptibility to errors during 3D modeling or texture fitting can lead to degraded results.

In this paper, we revisit and \textbf{re}-wire the powerful \textbf{S}tyle-based generator~\cite{karras2019style, Karras2019stylegan2} to generate faithful lip-\textbf{sync}ed videos and support virtual perform\textbf{er} creation in a concise yet effective framework called \textbf{ReSyncer}. With unified model training, various appealing properties 
including generalized and personalized lip-sync, speaking style transfer, live streaming, and face swapping can be achieved within one model.
Our key is to show \emph{how an easy reconfiguration empowers Style-based generator with various powerful audio-visual synchronization abilities by involving 3D facial dynamics}.
Specifically, we use only roughly fitted 3D facial meshes, instead of coefficients to bridge the audio and image domains. A principled \emph{Style-injected Lip-Sync Transformer (Style-SyncFormer)}  learns stylized point cloud displacements given template meshes and driving audios. 

We then perform easy but crucial modifications to Style-based generator with a simple encoder used in previous work~\cite{yang2021gan,xu2022styleswap,guan2023stylesync}.
Concretely, while lip-sync studies~\cite{prajwal2020lip,sun2022masked,cheng2022videoretalking,guan2023stylesync, wang2023facecomposer} mask out the mouth area of the target frame, we overlay reconstructed mesh onto it as strong spatial guidance. Then we compliment the textural and identity information with an attached reference frame, which serves as the source of both the noise and style-related space. 
With generalized training on massive data, the powerful generator focuses on recovering facial details from the roughly predicted lip-synced white-colored meshes, leading to high-fidelity and stable lip-sync with higher quality.

Additionally, our framework design goes beyond existing solutions by uniformly addressing a broader range of needs in virtual performer creation. These needs include, for instance, preserving personal styles for specific users or changing face identities for privacy protection and diverse character generation. 
Particularly, our method directly supports few-shot few-step personalized fine-tuning. Moreover, our framework supports \emph{syncing (swapping)} different identities on the same target video with additional training strategies. High-fidelity face-swapping and lip-syncing can hence be simultaneously satisfied.

Our contributions can be summarized as follows: 
\textbf{1)} We propose the \textbf{ReSyncer} framework, which shows the powerfulness of Style-based generator on syncing audio-visual facial information by involving 3D facial meshes with a simple reconfiguration.  
\textbf{2)} We propose the Style-SyncFormer which learns stylized 3D facial dynamics with simple Transformer blocks, enabling generalized 3D facial animation. 
\textbf{3)} Extensive experiments show that our framework not only enables lip-syncing of higher stability and quality, but also supports various intriguing properties that are essential for creating virtual performers, including fast personalized fine-tuning, video-driven lip-syncing, the transfer of speaking styles, and even face swapping. To the best of our knowledge, we are one of the first methods that support state-of-the-art \textbf{lip-syncing} and \textbf{face-swapping} with one unified model.

%% file: sec/2_related.tex
\section{Related Work}
\label{sec:related}

\subsection{Audio-Driven Facial Animation}

\noindent\textbf{Talking Head Generation.} A great number of studies focus on predicting the whole human head movements according to audios~\cite{suwajanakorn2017synthesizing,jamaludin2019you,chen2018lip, chen2019hierarchical, zhou2021pose,kaisiyuan2020mead, guo2021adnerf, ji2022eamm,liang2022expressive,ma2023styletalk,ma2023talkclip,wang2022pdfgc,wang2021audio2head,yu2022thpad,zhang2023sadtalker,qiu2024relitalk}. More recently, one-shot-based methods adopt representations from 3D, videos, and even texts for more controllable results. While other approaches~\cite{guo2021adnerf, liu2022semantic, shen2022dfrf, tang2022real, yao2022dfa, ye2023geneface} attempt to integrate speech audios into the neural radiance field for personalized audio-to-3D modeling. However, these methods usually fail to achieve stable lip-sync results and show terrible robustness performance when using audio in the wild. Moreover, the form of talking head generation cannot be applied to edit arbitrary video.

\noindent\textbf{Lip-Syncing-based Facial Editing.} Apart from generating the entire head, certain studies~\cite{prajwal2020lip, thies2020neural, song2018talking, park2022synctalkface, sun2022masked, cheng2022videoretalking, guan2023stylesync,wu2023speech2lip} choose to directly edit the mouth shape according to the input audios. Specifically, \cite{prajwal2020lip} relies on a well-trained lip-sync discriminator to produce highly synchronized animations. \cite{sun2022masked} introduces an audio-visual cross-modality framework based on transformer~\cite{chang2022maskgit}. Particularly, \cite{guan2023stylesync} builds a person-agnostic lip-sync framework by StyleGAN2~\cite{Karras2019stylegan2}, which also supports personalized optimization. Similarly, \cite{cheng2022videoretalking} also provides identity refinement for the final animation results. Although they have shown accurate lip-sync and comparably high-fidelity results, visible artifacts may still exist due to the latent contradiction between textural and spatial information.
Few studies~\cite{stypulkowski2023diffused, shen2023difftalk} built upon diffusion models~\cite{rombach2022high} are presented recently. Our ReSyncer generates high-fidelity lip-sync videos with high stability and efficiency.

\noindent\textbf{Audio-Driven 3D Facial Animation.} The first stage of our work shares a similar setting with methods that generate audio-driven 3D facial animation~\cite{cudeiro2019capture,richard2021meshtalk,fan2022faceformer,Xing_2023_CVPR,thambiraja2023imitator}. Most of them are performed on 3D scan data to predict the point clouds' dynamics. They have explored Transformers~\cite{vaswani2017attention} and codebooks~\cite{van2017neural} with different settings on this task. We adopt a highly efficient form to use Transformer in our work.

\subsection{Face Swapping}
To transfer the identity of one person to the face of another person is a long-standing problem in computer vision. In recent years, GAN~\cite{goodfellow2014generative} has played an important role in this task.
RSGAN\cite{natsume1804rsgan} and FSNet~\cite{natsume2019fsnet} encode the facial region and the non-facial region to different latent codes and perform swapping by replacing these latent representations. 
\cite{zhu2021one} proposes a face-swapping method MegaFS built upon the GAN-inversion technique with StyleGAN2, however, the results are not temporal consistent. Similarly, \cite{xu2022high, xu2022styleswap} make their attempts to apply the strong prior learned by StyleGAN2~\cite{Karras2019stylegan2} to generate high-resolution swapped faces.
Recently, \cite{li20233d} takes inspiration from 3D generative model~\cite{chan2022efficient} to capture fine-grained details of face shape and strengthen the robustness under large pose variance by extending the face swapping task into 3D latent space.
However, they fail to produce results with accurate poses and expressions when balancing identity similarity and attribution preservation of the target frame.

Different from these methods, ReSyncer takes advantage of intermediate guidance from 3D mesh to synthesize accurate pose and expression. Moreover, we achieve face-swapping in a compatible with accurate lip-syncing, which allows us to further satisfy more imaginative demands in creating virtual presenters and performers. 

%% file: sec/3_method.tex
\setlength{\abovecaptionskip}{4pt}
\begin{figure*}[!t]
\centering
\includegraphics[width=\linewidth]{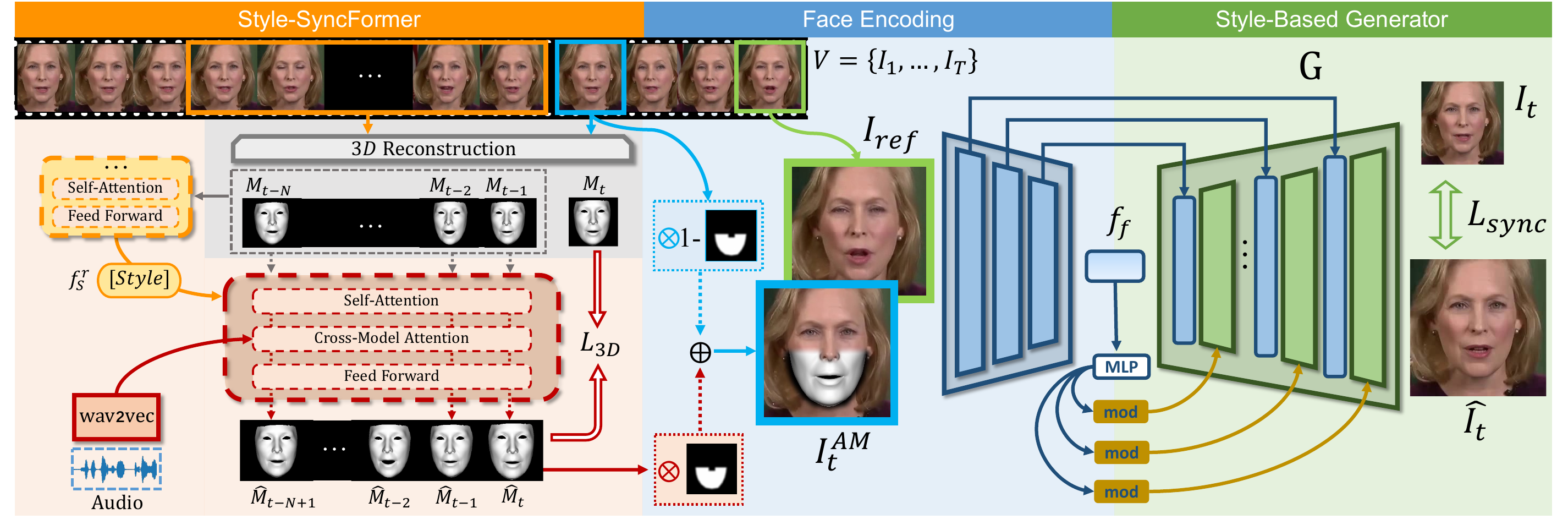}
\caption{\textbf{The ReSyncer Framework.} In the first stage, the \textbf{Style-SyncFormer} takes the style template and audio input to predict 3D facial dynamics. Then the predicted mesh overlays on the target frame to provide strong spatial guidance. The Style-based generator \text{G} processes the overlay and reference frames to produce the final result.}
\label{fig:pipeline}
\end{figure*}

\section{Methodology}
\label{sec:method}
In this section, we introduce the details of our ReSyncer framework as depicted in Fig.~\ref{fig:pipeline}. It consists of two stages which are all simply designed and easy to implement. In Sec.~\ref{sec:3.1} we illustrate our design of the style-injected lip-sync Transformer (Style-SyncFormer) which predicts 3D facial dynamics. 
Then we show how to reconfigure the Style-based generator to support high-fidelity generalized and personalized lip-sync in Sec.~\ref{sec:3.2}. Finally, we illustrate our unified design naturally supports face-swapping training in Sec.~\ref{sec:3.3}.

\noindent\textbf{Task Formulation.} The task of generalized lip-sync is normally trained on massive in-the-wild videos. Specifically, a training data consists of video clip $\textbf{V} = \{I_1, \dots, I_T\}$ with $T$ frames and the accompanying (processed) audio clip features $\textbf{a} = \{a_1, \dots, a_T\}$. 
The lower half of the target frame $I_t$ is masked out by a mask $A$ to form $I^A_t = (\textbf{1} - A) * I_t$ as~\cite{prajwal2020lip,sun2022masked,cheng2022videoretalking,guan2023stylesync}.
The normal training objective is to recover the original frame given the audio clip $a_t$ and a reference frame $I_{ref}$. 
However, the final results might be interfered with by the structure of the reference frame when the estimated animation significantly differs from the reference.

In this paper, we involve roughly predicted 3D facial meshes as intermediate representations. For every frame in the training video, we reconstruct the 3D facial meshes $\textbf{M} = \{M_1, \dots, M_T\}$. Thus our framework can be divided into the stages of predicting 3D facial dynamics from audio and the recovery of frames $I_t$ with the assistance of $M_t$.

\subsection{Style-Injected Lip-Sync Transformer}
\label{sec:3.1}
In the first stage, we propose the Style-SyncFormer that takes audio features $\textbf{a}_{t_0:t_1} = \{a_{t_0}, \dots, a_{t_1}\}$ extracted from pretrained Wav2Vec2~\cite{baevski2020wav2vec} to predict 3D meshes $\textbf{M}_{t_0:t_1} = \{M_{t_0}, \dots, M_{t_1}\}$.
There are $N$ consecutive frames between $t_0$ and $t_1$.
Previous work~\cite{fan2022faceformer} has explored abilities to learn 3D dynamics but it handles only a fixed number of identities.
In contrast, the proposed Style-SyncFormer facilitates zero-shot prediction of 3D dynamics while preserving the original speaking style or transferring it from another person.

\noindent\textbf{Data Processing.} During training, all pose information is eliminated from the meshes and aligned to the canonical space.
Instead of directly learning all vertices' positions, we learn the displacements compared to a template mesh $\bar{M}$, the mean of all meshes from this person without expression coefficients. The training target becomes predicting the displacements $\Delta \textbf{M}_{t_0:t_1} = \{\Delta M_{t_0}, \dots\, \Delta M_{t_1}\}$ between the template and each reconstructed result.

\noindent\textbf{Network Design.} The main network of Style-SyncFormer is built based on only one Transformer~\cite{vaswani2017attention} encoder block for style information encoding and one decoder block for mesh prediction, where we adopt the periodic positional encoding and biased causal multi-head attention from FaceFormer~\cite{fan2022faceformer}.

During training, each shifted-right displacement $\Delta M_t$ is mapped to embedding $d^m_{t}$ and sent into the positional encoding to serve as tokens $f^m_{t}$.
Previous studies~\cite{fan2022faceformer,richard2021meshtalk} learn the 3D dynamics of only limited identities with a simple one-hot speaker identity embedding~\cite{cudeiro2019capture,fan2022faceformer}. 
Differently, we encode the information of a total of $T_S = 90$ consecutive 
$\Delta M_{ref}$ as speaking style reference. The style information is embedded as a feature $f^r_{S}$ from $\Delta M_{ref}$ using an additional Transformer encoder block.
After that, we inject the speaking style $f^r_{S}$ by adding it with mesh tokens $f^m_{t}$.
Note that when replacing $\Delta M_{ref}$ with mesh sequences of another identity, speaking style transfer can be achieved.

In the decoder block, the input tokens are sent into multi-head self-attention, cross-attention, and feed-forward layers. The audio features are further encoded as embeddings $f^a_t$ and leveraged in the cross-attentions. The final output is transformed to $\Delta \hat{M}_{t_0:t_1} = \{\Delta \hat{M}_{t_0}, \dots, \Delta \hat{M}_{t_1} \}$.

\noindent\textbf{Loss Function.} The training objectives can be written as: 
\begin{align}
\mathcal{L}_{3D}  =  \sum_{t=t_0+1}^{t_1} \|(\Delta \hat{M}_t - \Delta \hat{M}_{t-1}) - (\Delta {M}_t - \Delta {M}_{t-1})\|_2 \nonumber  + \lambda_1 \sum_{t=t_0}^{t_1} \| \Delta \hat{M}_t - \Delta M_t \|_2,
\end{align}
where the first term is the temporal consistency loss~\cite{thies2020neural}.

The training is performed in a teacher-forcing manner while the network works autoregressively during inference by using previously predicted results. The displacements are finally transferred to 3D meshes $\hat{\textbf{M}}_{t_0:t_1}$ in the canonical space by adding the template mesh $\bar{M}$. 

\subsection{Rewiring Style-based Generator}
\label{sec:3.2}
In the second stage, we rewire the widely used Style-based generator~\cite{Karras2019stylegan2} to faithfully transfer 3D meshes $\textbf{M}$ to high-fidelity facial frames $\hat{\textbf{I}}$ in a simple manner.
Previous methods~\cite{thies2020neural,chen2020talking} resort to accurately textured rendering, which is inefficient from the perspective of implementation and training. Our goal is to find an efficient and effective way.
As a result, we leverage the basic white-colored rendering as spatial guidance, which can be easily achieved without textural mapping.

\noindent\textbf{Network Design.} 
We resort to 
Style-based generator with noise insertion scheme, which is easy to implement and specializes well in face-related tasks~\cite{yang2021gan,xu2022styleswap,guan2023stylesync,luo2022styleface}.
Specifically, we project mesh $M_t$ on masked target frame  $I^A_t$'s domain to $M^p_t$ and overlay to form a 3D-guided image 
$I^{AM}_t = I^A_t + A * M^p_t$. 
A reference frame $I_{ref}$ is attached for providing textural information in the spatial form. The 6-channel input 
$[I^{AM}_t, I_{ref}]$ is encoded to feature maps $\textbf{F}_f = \{f_f, F_1, \dots, F_L\}$ that match the sizes of Style-based generator's each stage and $f_f$ being a vector. Then we follow previous studies~\cite{yang2021gan,xu2022styleswap,guan2023stylesync} and insert $\textbf{F}_f$ into the generator $\text{G}$. As $I^{AM}_t$ directly provides the 3D mouth shape information, the reference frame can hardly influence the prediction result as long as it is not the target frame itself. Thus we randomly sample $I_{ref}$ within the nearest 30 frames of $I_t$.

A number of previous lip-sync studies~\cite{zhou2021pose,guan2023stylesync,liang2022expressive,wang2022pdfgc,yu2022thpad} rely on encoding audio information into the $\mathcal{W}$ space for animation. However, with explicit guidance, the driving information would play a less important role in the $\mathcal{W}$ space. Meanwhile, the face-swapping task illustrates that $\mathcal{W}$ space embedding benefits identity information~\cite{zhu2021one,xu2022styleswap,luo2022styleface} which we intend to enhance. To avoid involving extra network design and inference procedures, we directly send the encoded $f_f$ from $[I^{AM}_t, I_{ref}]$ to the $\mathcal{W}$, so that the reference frame provides more facial detail while $I^{AM}_t$ also passes dynamic information.

\noindent\textbf{Loss Functions.} To sum up, in the second stage, the generator takes the output only of a single facial encoder and predicts the target frame itself as $\hat{I}_t$. The training objective is much simpler with the 3D guidance. We only adopt the same loss functions as Pix2pix-HD~\cite{wang2018high}, which are the $L_1$, VGG, and adversarial loss empowered with StyleGAN2's discriminator.
\begin{align}
    \mathcal{L}_{rec} = \|\hat{I}_t - I_t\|_1 +   \lambda_2 \sum_{m=1}^{N_{vgg}}\|\text{VGG}_m(\hat{I}_t) - \text{VGG}_{m}(I_t) \|_1,
\end{align}
\begin{align}
    \mathcal{L}_{adv} =  \underset{\text{G}}{\text{min}}\underset{\text{D}}{\text{max}}(\mathbb{E}_{I_{t}}[\log \text{D}(I_{t})]  
     + \mathbb{E}_{\hat{I}_t}[\log (1 - \text{D}(\hat{I}_t))]).
\end{align}
As the rendered meshes provide much clearer guidance than audio, the generator can be easily trained without task-oriented losses or specific tuning.

During inference, we overlay the predicted mesh $\hat{M}^p_t$ from the first stage to form the input $[\hat{I}^{AM}_t, I_{ref}]$. Though the predicted $\hat{M}^p_t$ cannot perfectly match the reconstructed results, we rely on the powerful generator to compensate for such disturbance.

\noindent\textbf{Video-Driven Lip-Sync.} The generator only depends on mesh information, thus we can transfer the displacements of meshes $\Delta \textbf{M}'$ of another person to the target template $\bar{M}$ so that video-driven lip-sync can be naturally achieved.

\noindent\textbf{Personalized Fine-tuning.} 
Given that syncing information is derived from the 3D facial mesh, the generator in the second stage will not overfit the limited lip patterns present in given clips. This masks our method full-parameter fine-tunable on only seconds of videos.

\subsection{Face-Swapping}
\label{sec:3.3}
We now illustrate the power of our comprehensive framework to directly achieve face-swapping and lip-sync with unified training.
Our design of leveraging the $\mathcal{W}$ space for enhancing facial generation considers the similar setting used in previous face swapping studies~\cite{xu2022styleswap,xu2022mobilefaceswap,luo2022styleface}. Moreover, the identity information of the audio-predicted mesh $\hat{M}_t$ can be easily transferred by sending different template meshes into the style encoder of the Style-SyncFormer. So that the shape of facial organs can be seamlessly swapped to another person.

\begin{figure}[t]
\centering
\includegraphics[width=.7\linewidth]{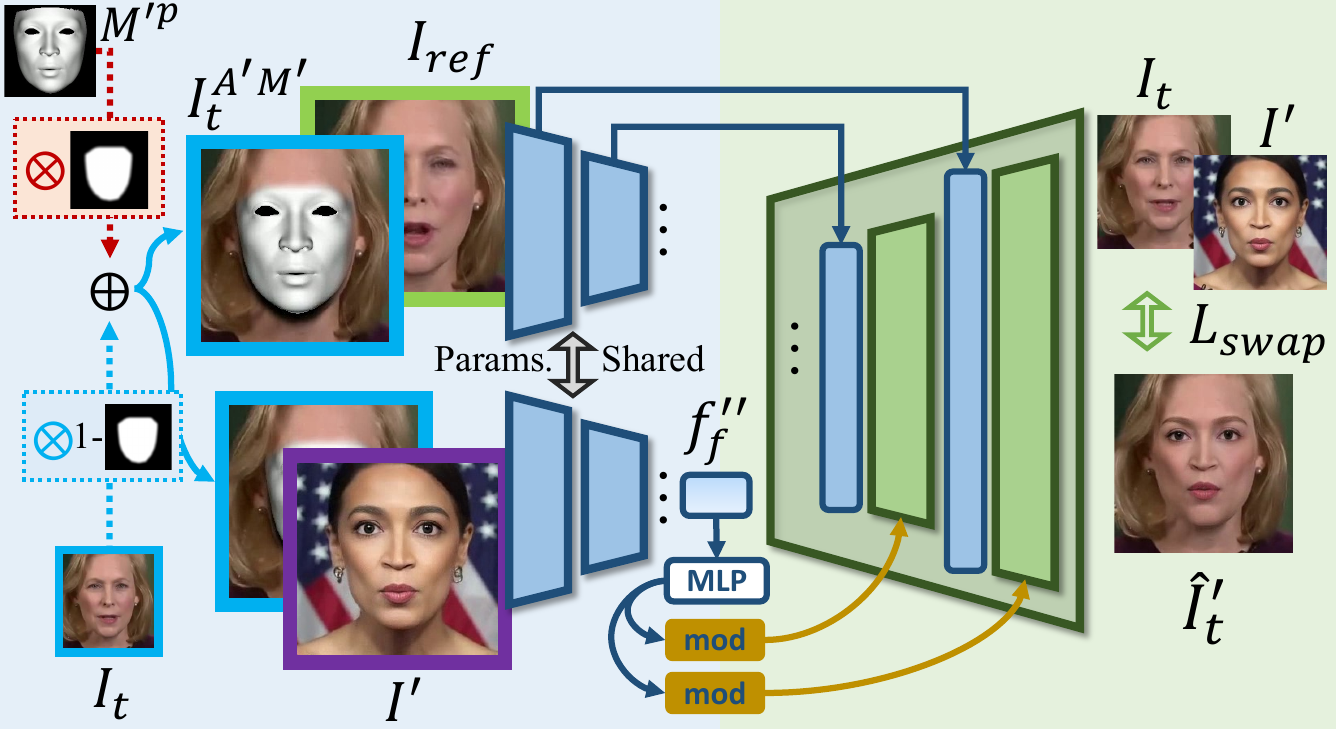}
\caption{\textbf{Face-Swapping Pipeline}. With input data reconfiguration and additional training losses, we achieve lip-syncing and face-swapping simultaneously.}
\label{fig:faceswapping}
\end{figure}

\noindent\textbf{The Face-Swapping Data Stream.} The pipeline is depicted in Fig.~\ref{fig:faceswapping}. 
We first enlarge the masking area $A$ to $A'$ that contains the whole facial part including eyes and noses, which is more suitable for the concerned task. Then we involve a frame $I'$ of another identity and its corresponding mesh $M'$ to provide additional information. We project $M'$ with the pose of $M^p_t$ to ${M'^p}$ and overlay it on $I_t$ to form 
$I^{A'M'}_t = (\textbf{1} - A') * I_t + A' * {M'^p}$.

The face-swapping training consists of two parts, updating only the Style-based generator. The {first} is the same as Sec.~\ref{sec:3.2}, which reconstructs $I_t$ solely from $\{I^{AM}_t, I_{ref}\}$. The {second} part intends to create an image $\hat{I'_t}$, which has $I'$'s identity on $I_t$'s head with $I_t$'s original tone. This requires the generator G to take $\textbf{F}'_f = \{f''_f, F'_1, \dots, F'_L\}$ encoded from different sources, including $f''_f$  encoded from  $\{I^{A'M'}_t, I'\}$ for ID injection in the $\mathcal{W}$ space and $\{F'_1, \dots, F'_L\}$ from $\{I^{A'M'}_t, I_{ref}\}$ for expressing facial attributes in the noise space.

\noindent\textbf{Training Objectives.} Under the cross-identity training setting, the commonly used identity cosine distance loss $\mathcal{L}_{id} = 1 - \cos(E_{id}(\hat{I}^{'}_t), E_{id}(I'))$ where $\text{E}_{id}$ is the encoder of ArcFace~\cite{deng2019arcface} and feature matching loss with the original $I_t$~\cite{chen2020simswap,xu2022region,xu2022styleswap} in the face-swapping task are leveraged.

Note that after the network is trained in the face-swapping style, which costs more time, high-quality face-swapping and lip-sync abilities can be achieved within one unified model (with the same set of weights). To the best of our knowledge, we are one of the first methods that unifies the two tasks within one model, which is efficient and beneficial to the creation of virtual performers.

%% file: sec/4_exp.tex
\setlength{\abovecaptionskip}{4pt}
\begin{figure*}[!t]
\centering
\includegraphics[width=\linewidth]{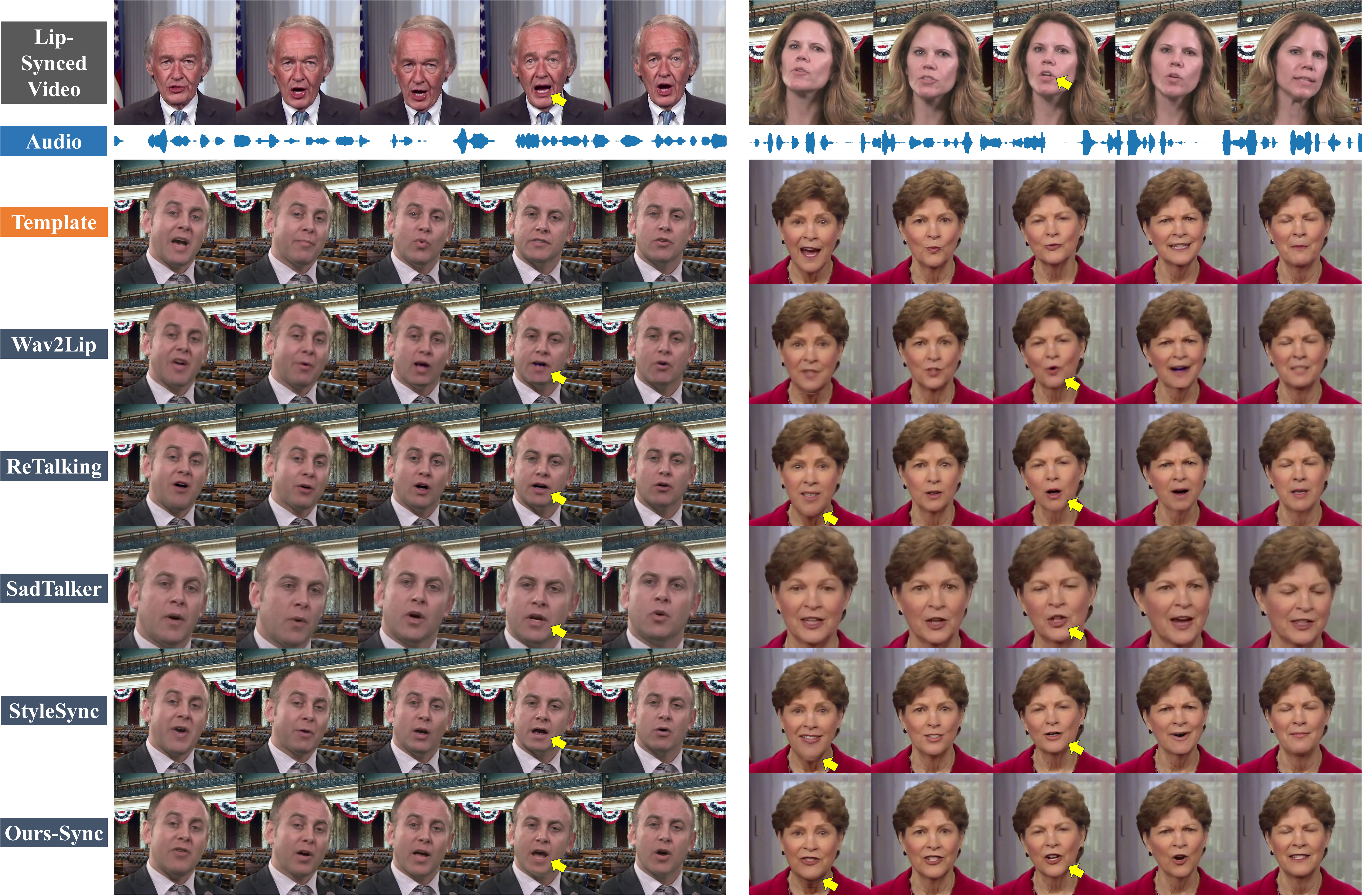}
\caption{
\textbf{Qualitative Cross-Sync Results}. 
The top row shows the lip-synced videos of the driving audio. Generation results based on the ``Template'' row should have the same lip shape as the ``Lip-Synced Video'' in the first row.
}
\label{fig:cmp_sync}
\end{figure*}

\section{Experiments}
\label{sec:experiment}
Although lip-sync is our main concern,
we thoroughly conduct experiments from the perspectives of both lip-sync and face-swapping to show the effectiveness of our unified framework. For fair comparisons, we present results using only our generalized model, please refer to the appendix~\ref{sec:compare_personalized_method} for personalized experiments.

\noindent\textbf{Datasets.}
We train our model with two audio-visual datasets, HDTF~\cite{zhang2021flow} and VoxCeleb2~\cite{Chung18b}. Both two datasets provide high-quality videos from a variety of identities for lip-syncing and face-swapping training. 
Notably, we split HDTF into train and test sets without overlap. The test set comprises 100 10-second audio-video pairs.
Evaluations on lip-sync are conducted on the official test set of VoxCeleb2 and the introduced test set of HDTF.  
Face-swapping evaluation is performed on FaceForensics++ (FF++) \cite{roessler2019faceforensicspp}, which contains 1000 face videos. We use the high-quality version and adopt the official video-video pairing.
Apart from publicly available datasets, certain visual outcomes in our experiments are produced based on videos sourced from the Internet.

\noindent\textbf{Implementation Details.}
All videos are processed at a frame rate of 25FPS, and facial alignment is conducted based on pre-detected landmarks. 
Subsequently, all faces are cropped to a size of $256\times256$. 
The architecture of \text{G} is built upon the success of recent methods \cite{guan2023stylesync,xu2022styleswap,karras2020analyzing,yang2021gan}. 
During inference, we split videos into 100-frame consecutive chunks with a 20-frame overlap.
Face meshes are predicted in an auto-aggressive manner with the proposed Style-SyncFormer depending on all the previously predicted results within each chunk.
For face reconstruction, we train DeepFaceReconstruction~\cite{deng2019accurate} on the topology of both Hifi3DFace~\cite{hifi3dface2021tencentailab} and BFM. The main results are reported from the Hifi3DFace version but the choice of topology makes little difference.
Our models are trained on 4 Tesla A100 GPUs for 2 days with batch size 16. 
For more implementation details, please refer to the appendix~\ref{sec:supp_additional_implementation}.

\noindent\textbf{Comparison Methods.}
As for lip-sync evaluation,
We compare our method with four different methods including the classic Wav2Lip~\cite{prajwal2020lip}, as well as three recent state-of-the-art (SOTA) methods: ReTalking~\cite{cheng2022videoretalking}, SadTalker~\cite{zhang2023sadtalker}, and StyleSync~\cite{guan2023stylesync}. 
Note that SadTalker~\cite{zhang2023sadtalker} targets talking head generation but is well-known for its good quality.  All experiments are conducted on the generalized setting without fine-tuning, and the model without face-swapping training is leveraged.
Then we compare our method with four studies on face swapping. The counterparts consist of SimSwap~\cite{chen2020simswap}, InfoSwap~\cite{gao2021information}, StyleSwap~\cite{xu2022styleswap} and E4S~\cite{liu2023fine}.
Furthermore, we aim to evaluate whether combining two methods from the above two tasks can achieve comparable results with our full model despite the cost.
Thus we construct two potential counterparts by combining SOTA methods from both areas, denoted as ``SimSwap + ReTalking'' and ``StyleSwap + StyleSync''.

\begin{table*}[tb]
\centering
\setlength{\belowcaptionskip}{2pt}
\caption{{Quantitative results on HDTF and VoxCeleb2.} For LMD and $\Delta{\text{Sync}}$ the lower the better, and the higher the better for others.}
\resizebox{\linewidth}{!}{
\begin{tabular}{lcccccccc}
\toprule
\multirow{2}{*}{Method} & \multicolumn{4}{c}{HDTF}            & \multicolumn{4}{c}{VoxCeleb2}     \\ 
\cline{2-9} 
\noalign{\smallskip}
& SSIM $\uparrow$ & PSNR $\uparrow$ & LMD $\downarrow$ & $\Delta{\text{Sync}}$ $\downarrow$ & SSIM $\uparrow$ & PSNR $\uparrow$ & LMD $\downarrow$ & $\Delta{\text{Sync}}$ $\downarrow$  \\ 
\hline
\noalign{\smallskip}
Wav2Lip~\cite{prajwal2020lip}  & \textbf{0.84} & 30.93 & 5.41 & 1.37 & 0.84 & 31.58 & 3.44 & 2.28 \\
ReTalking~\cite{cheng2022videoretalking}  & 0.81 & 31.08 & 6.06 & 0.74 & 0.83 & 31.64 & 4.42 & 1.50 \\
SadTalker~\cite{zhang2023sadtalker} & 0.77 & 30.86 & 6.76 & 3.62 & 0.79 & 31.36 & 5.03 & 3.03 \\
StyleSync~\cite{guan2023stylesync}  & 0.83 & 31.57 & 4.97 & 0.80 & \textbf{0.85} & 32.52 & 3.28 & 1.57 \\
\hline
\noalign{\smallskip}
Ours-Sync & \textbf{0.84} & \textbf{31.76} & \textbf{4.34} & \textbf{0.66} & \textbf{0.85} & \textbf{32.88} & \textbf{3.19} & \textbf{0.88} \\
\bottomrule
\end{tabular}
}
\label{tab:exp_cmp_self_sync}
\end{table*}

\begin{table*}[b] 
\begin{minipage}{0.54\textwidth}
\setlength{\belowcaptionskip}{2pt}
\centering
\caption{Quantitative results evaluated on FF++. 
}
\resizebox{\linewidth}{!}{
\begin{tabular}{lcccc}
\toprule
Method & ID Retrieval $\uparrow$ & ID Sim. $\uparrow$ & Pose Error $\downarrow$ & Exp. Error $\downarrow$ \\
\midrule
SimSwap~\cite{chen2020simswap}  & 93.07\% & 0.578 & {1.36} & \textbf{5.07} \\
InfoSwap~\cite{gao2021information} & 95.82\% &0.635 & 2.54 & 6.99 \\
StyleSwap~\cite{xu2022styleswap} & \textbf{97.05}\% & \textbf{0.677} & 1.56 & 5.28 \\
E4S~\cite{liu2023fine}  & 90.21\% & 0.495 & 2.78 & 11.52 \\
\hline
\noalign{\smallskip}
Ours-Swap  & \underline{96.70\%} & \underline{0.665} & \textbf{1.31} & \underline{5.13} \\
\bottomrule
\end{tabular}
}
\label{table:cmp_swap_metrics}
\end{minipage}
\begin{minipage}{0.44\textwidth}
\setlength{\belowcaptionskip}{2pt}
\centering
\caption{Quantitative ablations on VoxCeleb2.}
\resizebox{\linewidth}{!}{
\begin{tabular}{lcccc}
\toprule
Method       & SSIM $\uparrow$ & PSNR $\uparrow$  & LMD $\downarrow$ & $\Delta\text{Sync}$ $\downarrow$  \\ 
\midrule
w/o Mesh-Inject & 0.83 & 31.57 & 3.60 & 2.54 \\
w/o Mesh-Style & 0.83 & 31.80 & 3.29 & 1.02 \\
$T_S=50$ & \textbf{0.85} & 32.26 & 3.35 & 0.95 \\
$T_S=200$ & \textbf{0.85} & 32.69 & 3.21 & \textbf{0.88} \\
Ours-Sync & \textbf{0.85} & \textbf{32.88} & \textbf{3.19} & \textbf{0.88} \\
\bottomrule
\end{tabular}
}
\label{tab:sync_ablation}
\end{minipage}
\end{table*}

\subsection{Quantitative Comparisons}
\label{sec:Quantitative Comparisons}
Quantitative experiments of lip-sync are conducted following the self-reconstruction setting. We refrain from directly utilizing the ground truth frame as the reference. 
For face-swapping, we adhere to a protocol as previous studies \cite{xu2022styleswap, li2019faceshifter}, wherein we uniformly sample 10 frames from each video in FF++, resulting in a total of 10k images for evaluation.

\noindent\textbf{Evaluation Metrics.}
Both lip-sync and face-swapping are broadly studied, we follow previous works \cite{guan2023stylesync,xu2022styleswap,li2019faceshifter,zhou2021pose} to adopt several popular metrics.
For lip-sync, we use SSIM \cite{wang2004image} and PSNR for measuring visual similarity. Landmark distances around the mouth (LMD), and SyncNet’s confidence score \cite{chung2016out} are used for lip-sync quality. 
Particularly, the SyncNet score only indicates how well a video aligns with the learned SyncNet model, thus we use the absolute difference ($\Delta$Sync) between ground truth and prediction to assess whether results are close to the original video.
For face-swapping, we leverage a pretrained face-recognition model \cite{wang2018cosface} to measure the ID cosine similarity and ID retrieval scores. 
In addition, pose error and expression error are measured by two pretrained models\cite{ruiz2018fine,vemulapalli2019compact} following previous work~\cite{xu2022styleswap}.

\setlength{\abovecaptionskip}{4pt}
\begin{figure*}[!t]
\centering
\includegraphics[width=\linewidth]{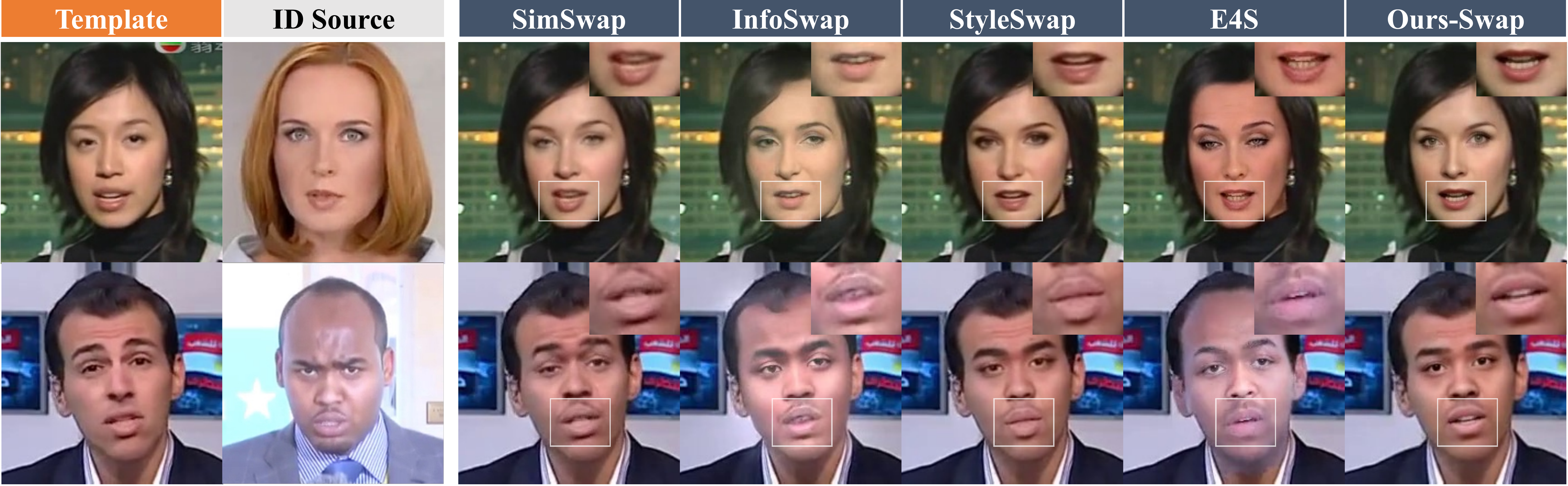}
\caption{\textbf{Qualitative Results of Face-Swap.} Identity-swapped results should preserve the expression and lip motion of templates.}
\label{fig:cmp_self_swap}
\end{figure*}

\noindent\textbf{Lip-Sync.}
The evaluation results are tabulated in Table~\ref{tab:exp_cmp_self_sync}. 
Our method outperforms previous works across all metrics, especially concerning lip-sync quality.
This is attributed to stylized 3D facial dynamics learned by Style-SyncFormer, which manages to recover the injected personalized speaking style of the target person in the generalized setting. In contrast, previous methods primarily rely on sync-loss \cite{prajwal2020lip} and tend to generate a dataset-averaged sync-style.

\noindent\textbf{Face-Swapping.}
Results evaluated on FF++ are listed in Table~\ref{table:cmp_swap_metrics}. Our model shows comparable performance on all the metrics and achieves the lowest pose error when compared with SOTA face-swap methods. Particularly, our performances on ID-related metrics are only slightly weaker than StyleSwap.

\subsection{Qualitative Comparisons}
\label{sec:Qualitative Comparisons}
Subjective evaluation plays a pivotal role in assessing the capabilities of generative models, especially in the context of videos. We strongly encourage readers to watch our \textbf{supplementary video} for a comprehensive understanding.
Our method also enjoys the capability to fast optimize for specific identities, resulting in superior outcomes. 
Please refer to the appendix~\ref{sec:compare_personalized_method} for the personalized results. 

\noindent\textbf{Lip-Sync.}
We demonstrate two cross-sync comparisons with SOTA methods in Fig.~\ref{fig:cmp_sync}.
The lip-synced results are conditioned on audio randomly sampled from the dataset. 
Wav2Lip has accurate lip shapes, but the mouth part is noticeably blurred.
SadTalker achieves better fidelity, but it fails to present natural dynamics.
VideoReTalking and StyleSync can generate precise and high-quality results; nevertheless, their results are affected by the opening jaw of the template as shown in the first column of the right figure. Moreover, they exhibit a deficiency in maintaining person-specific attributes around the mouth. Our method directly produces high-fidelity results with accurate lip motion and rich details of the mouth.

We present cross-sync results, showcasing the transferred speaking style in Fig.~\ref{fig:teaser} and Fig.~\ref{fig:show_trans}. It can be seen that our method not only accurately generates lip shapes aligned with the driven video but also successfully animates the target person with the speaking style of the source person in the driving video. 
In addition, our method inherently supports video-driven lip-syncing as introduced in Sec.~\ref{sec:3.2}. The comparison is demonstrated in Fig.~\ref{fig:cmp_video_driven}, it shows that person-specific attributes can be preserved better with Style-SyncFormer guidance.

\begin{figure}
\centering
\includegraphics[width=.8\linewidth]{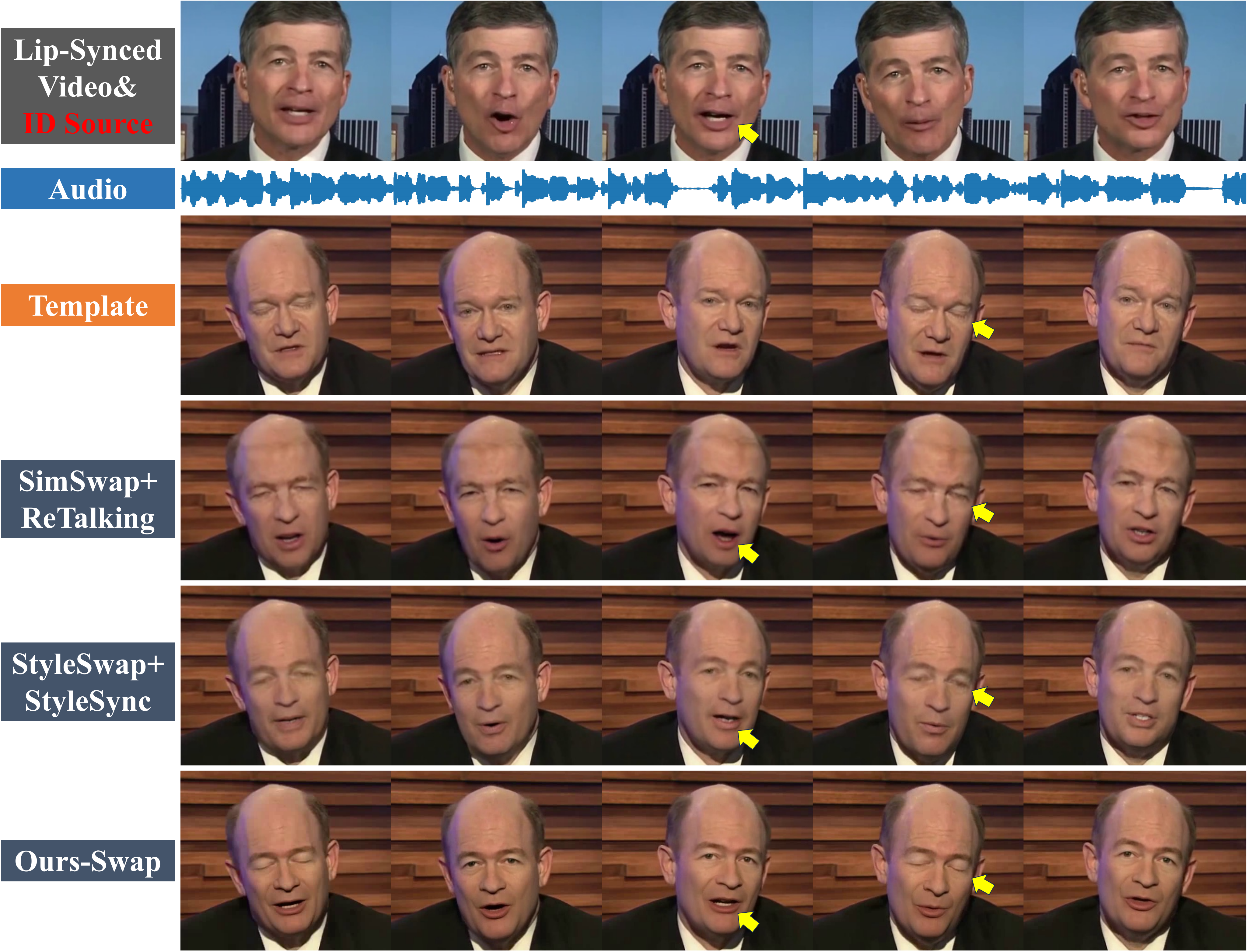}
\caption{\textbf{Results of Face-Swapped Lip-Sync.} The ID-swapped results are driven by the given audio. We compare with the combinations of two SOTA methods in lip-sync and face-swapping. Results generated by our method preserve details with better fidelity and simultaneously maintain a similar speaking style to the source.
}
\label{fig:cmp_cross_swap}
\end{figure}

\noindent\textbf{Face-Swapping.}
In Fig.~\ref{fig:cmp_self_swap}, we illustrate two face-swapped results for comparison. The generated results in this context should preserve the original expression and lip motion of the template.
From the figure, InfoSwap fails to keep a natural brightness across frames. 
E4S effectively modifies the entire head region but fails to integrate the head part into the background seamlessly. 
SimSwap and StyleSwap produce good results by preserving precise expression and identity attributes. However, an excessive emphasis on identity feature matching causes them to falter in generating details around certain areas like eyes and mouth.
In contrast, our method not only achieves effective identity transfer but also preserves rich details across the entire face.
In addition, the videos generated by our method exhibit superior temporal consistency, especially in the mouth region. Please find more results in the supplementary video.

Furthermore, our method is capable of achieving lip-synced face-swapping conditioned on any audio and identity in a single forward process.
We compare our method with two potential baselines ``SimSwap + ReTalking'' and ``StyleSwap + StyleSync'' in Fig.~\ref{fig:cmp_cross_swap}. 
Results generated by our method preserve details with better fidelity and simultaneously maintain a similar speaking style to the source.

\noindent\textbf{User Study.}
We conducted a user study with 30 participants. 
We adopt the commonly used Mean Opinion Scores (MOS) rating protocol. 
All users give their

\begin{table}[tb]
\centering
\caption{{User study of generalized lip-sync.} The scores are ranged from 1 (worst) to 5 (best).}
\resizebox{.8\linewidth}{!}{
\begin{tabular}{cccccc}
\toprule
MOS on & {Wav2Lip} & {ReTalking} & {SadTalker} & {StyleSync} & {{Ours-Sync}} 
\\
\midrule
Lip-Sync Quality  & 3.43 & 3.09 & 2.98 & 3.46 & \textbf{4.57} \\
Generation Quality & 1.48 & 2.83 & 3.13 & 3.63 & \textbf{4.67} \\
Video Realness     & 1.59 & 2.41 & 2.87 & 3.11 & \textbf{4.54}  \\
\bottomrule
\end{tabular}
}
\label{table:sync_user_study}
\end{table}

\noindent ratings, ranging from 1 to 5. On generalized lip-sync, they are asked to evaluate: 
a) the quality of lip-sync. 
b) visual quality of the generated video, and 
c) whether this video looks real. 
The results averaged from 10 videos are grouped into Table~\ref{table:sync_user_study}.
From the table, our method outperforms the four counterparts in all three aspects by clear margins, underscoring the power of our method and the substantial improvement in visual quality and realism over other existing methods.
The experiments are similarly conducted with face-swapping and personalized lip-sync, please find the results in the appendix~\ref{sec:supp_additional_exp}.

\subsection{Ablation Study}

\begin{figure}[t]
\centering
\includegraphics[width=.8\linewidth]{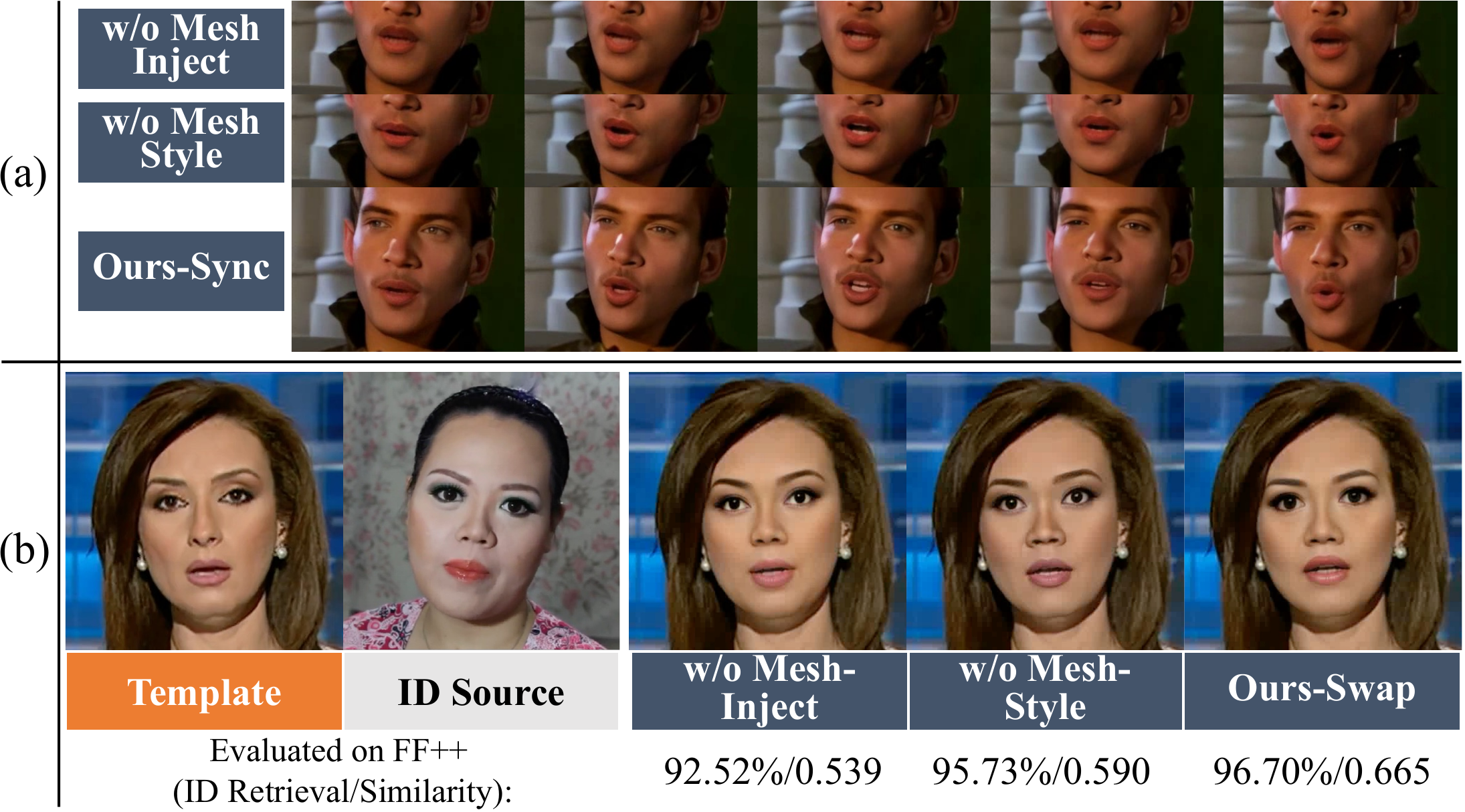}
\caption{
\textbf{Ablations.}
(a) 3D facial mesh provides detailed spatial guidance, empowering superior lip-syncing.
(b) The mesh with shapes of facial organs also enhances ID transfer in face-swapping.
}
\label{fig:ablations}
\end{figure}

\noindent\textbf{Lip-Sync.}
Here we evaluate four different designs on VoxCeleb2~\cite{Chung18b} including: 
1)``w/o Mesh-Inject'':  we mask-out the 3D mesh guidance from the encoded feature maps $\{F_1, \dots, F_L\}$, thus it cannot provide detailed spatial information during generation;
2) ``w/o Mesh-Style'': 3D mesh content is removed in $I^{AM}_t$ to get a $f_f$, feature vector that feed into the $\mathcal{W}$ space contains only the background and reference information; 
3) ``$T_S=50$'': we reduce the length ($T_S$ in Sec.~\ref{sec:3.1}) of the speaking-style reference from 90 to 50; 
4) ``$T_S=200$'': the length is extended to 200. 
The results are presented in Table~\ref{tab:sync_ablation}. 
Comparisons between the first two rows and ``Ours-Sync'' reveal the vital role played by mesh guidance in the generation process, where the sync quality is degraded to certain degrees, as indicated by LMD and $\Delta$Sync. In addition, the visual quality is also harmed as shown in Fig.~\ref{fig:ablations} (a). In the bottom three rows of Table~\ref{tab:sync_ablation}, we compare the performance of different sequence lengths in the speaking-style encoding of Style-SyncFormer. The results indicate that a sequence of 90 frames ($\sim$4s of video) leads to good sync quality, and a further extension of the length does not enhance its performance.

\noindent\textbf{Face-Swapping.}
In face-swapping, the predicted 3D facial mesh also provides detailed facial attributes. We provide an intuitive comparison in Fig.~\ref{fig:ablations} (b). 
When the mesh guidance is removed from the encoded feature map (``w/o Mesh-Inject''), we observe a noteworthy degradation in facial dynamics, particularly in lip motion. Please refer to the supplementary video for the comparison. Simultaneously, without the guidance of mesh information, the ID-related metrics of face-swapping are also affected, as indicated at the bottom of Fig.~\ref{fig:ablations} (b) and the same phenomenon applies to ``w/o Mesh-Style''. 

%% file: sec/5_conclusion.tex
\section{Discussion and Conclusion}
\label{sec:conclusion}

\noindent\textbf{Ethical Considerations.}
Our method possesses the capability to create fabricated talks and speeches, even with swapped identities, presenting a potential for malicious use. We will restrict the distribution of our models and the generated results exclusively to research purposes.

\noindent\textbf{Conclusion.}
We highlight several important properties of our \textbf{ReSyncer} framework: 
\textbf{1)} Our easy reconfiguration further unveils the potential of widely studied structures so that high-quality generalized lip-sync results with mesh representation can be achieved.
\textbf{2)} Our framework is designed to adopt external identity information, thus we achieve face-swapping comparable to the state of the arts while keeping the lip-sync ability within one unified model. 
\textbf{3)} Our network supports speaking style transfer, video-driven facial animation, and can be applied to real-time live streamers. These properties complementarily satisfy various needs of virtual performer creation under different circumstances.

\noindent\textbf{Acknowledgments.}
This work is in part supported by National Natural Science Foundation of China with No. 62394322, Beijing Natural Science Foundation with No. L222024, as well as the Beijing National Research Center for Information Science and Technology (BNRist) key projects.
This study is also supported by the Ministry of Education, Singapore, under its MOE AcRF Tier 2 (MOET2EP20221- 0012), NTU NAP, and under the RIE2020 Industry Alignment Fund – Industry Collaboration Projects (IAF-ICP) Funding Initiative, as well as cash and in-kind contribution from the industry partner(s).

%% file: sec/99_appendix.tex
\appendix
\section{Appendix}

\subsection{More Implementation Details}
\label{sec:supp_additional_implementation}
We introduce more implementation details of our ReSyncer framework.

\noindent\textbf{Architectures.}
As written in the main paper, we leverage only one Transformer encoder block for style information encoding and one Transformer decoder block for prediction.
Stacking more layers can only slightly improve the performance, we therefore adopt a light configuration for faster optimization and inference. A total of 100 128-dimensional tokens are sent into the decoder during training.
For audio encoding, we leverage a pretrained Wav2Vec2~\cite{baevski2020wav2vec} model and fix the temporal convolutional layers.
In the second stage of our method, we construct the generator \text{G} with a total of 14 style-convolution layers. 
The 3D facial meshes $\textbf{M} = \{M_1, \dots, M_T\}$ are reconstructed using a pretrained DeepFaceReconstruction model~\cite{deng2019accurate}. We adapted the implementation from~\cite{deng2019accurate} and re-train the model to adapt the topology of Hifi3DFace~\cite{hifi3dface2021tencentailab} with a similar performance of DeepFaceReconstruction itself. The change in topology does not affect our pipeline.

\noindent\textbf{Reference Image.}
The reference image $I_{ref}$ in our second stage is chosen with different strategies.
\textbf{1)} During \textbf{training}, the reference image is randomly chosen from the 30 closest frames to the target within the same video.
\textbf{2) Same-ID evaluation}. We form a reference sequence by shifting the background video by 75 frames (3s) to avoid information leakage.
\textbf{3) Cross-ID video production}. There is no information leakage when audio comes from another clip. The reference frame is selected as the target.

\noindent\textbf{Generalized Optimization.}
The whole model is trained using the Adam optimizer with a learning rate of $10^{-3}$. 
$\lambda_1$ and $\lambda_2$ are empirically set to 1, and 0.1, respectively.
Note during the initial stages of training, the first stage may not generate a facial mesh conducive to guiding our generator effectively. Consequently, we substitute the output of Style-SyncFormer with the reconstructed facial mesh from the pretrained DeepFaceReconstruction model until a certain convergence is achieved.

\noindent\textbf{Personalized Optimization.}
Given that syncing information is derived from the 3D facial mesh, the generator in the second stage will not overfit the limited lip patterns present in the given clips. This masks our method full-parameter fine-tunable on only seconds of videos. 
Meanwhile, as our method is built upon the Style-based generator, we can directly take advantage of the personalized fine-tuning protocols of~\cite{guan2023stylesync} with $\mathcal{W}^+$ space optimization to perform fast personalization.
We empirically found both $\mathcal{W}^+$ space finetuning and full-parameter tuning lead to similar results. Thus we directly apply full-parameter fine-tuning for video production in later personalized experiments. The hyperparameters and objective losses are kept the same as the generalized model.

\begin{figure}[t]
\centering
\includegraphics[width=\linewidth]{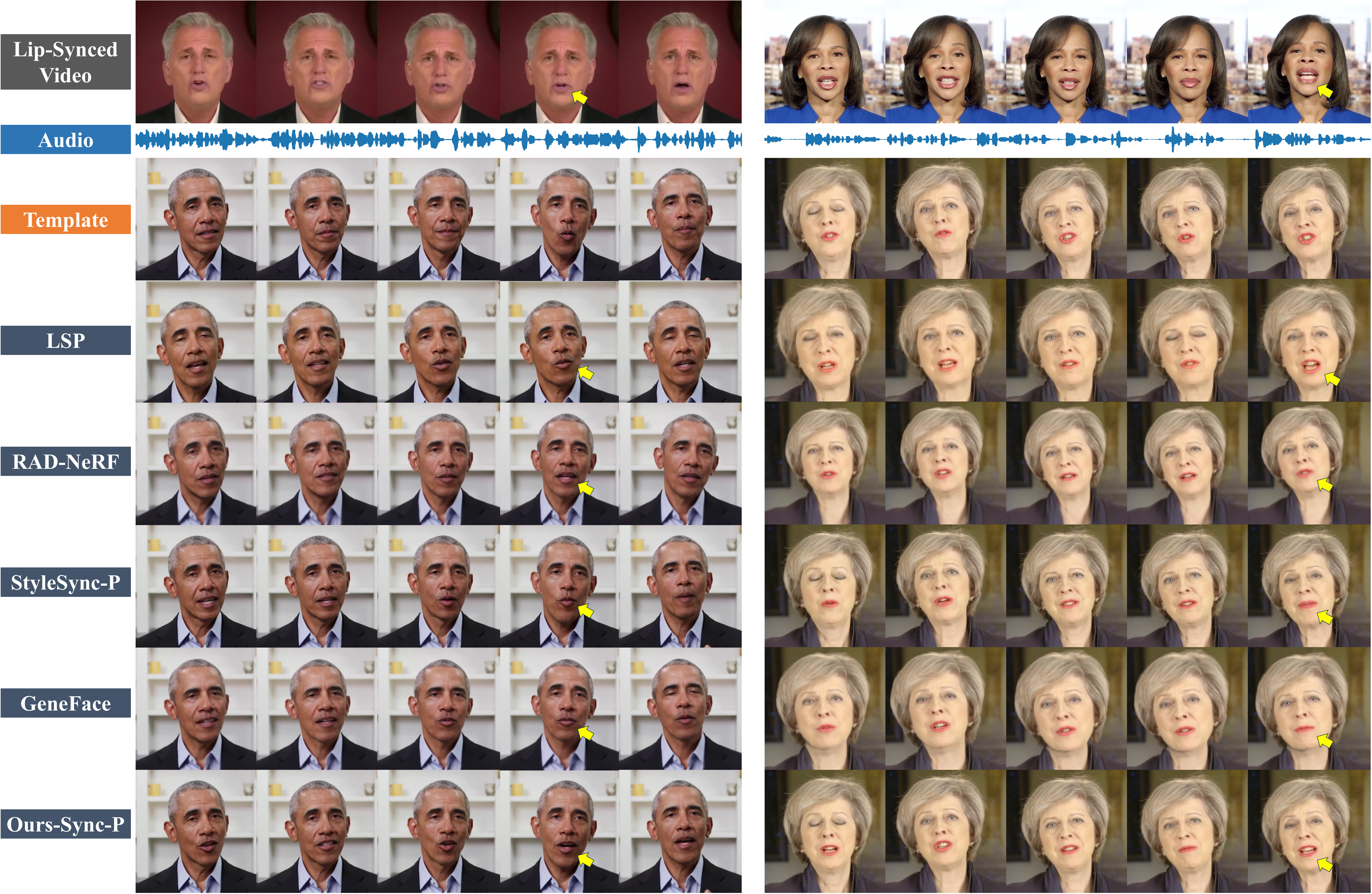}
\caption{\textbf{Qualitative Personalized Cross-Sync Results.} The top row shows the lip-synced videos of the driving audio. Generation results based on the ``Template'' row should have the same lip shape as the ``Lip-Synced Video'' in the first row.}
\label{fig:cmp_personalized_method}
\end{figure}
\begin{figure}[t]
\centering
\includegraphics[width=\linewidth]{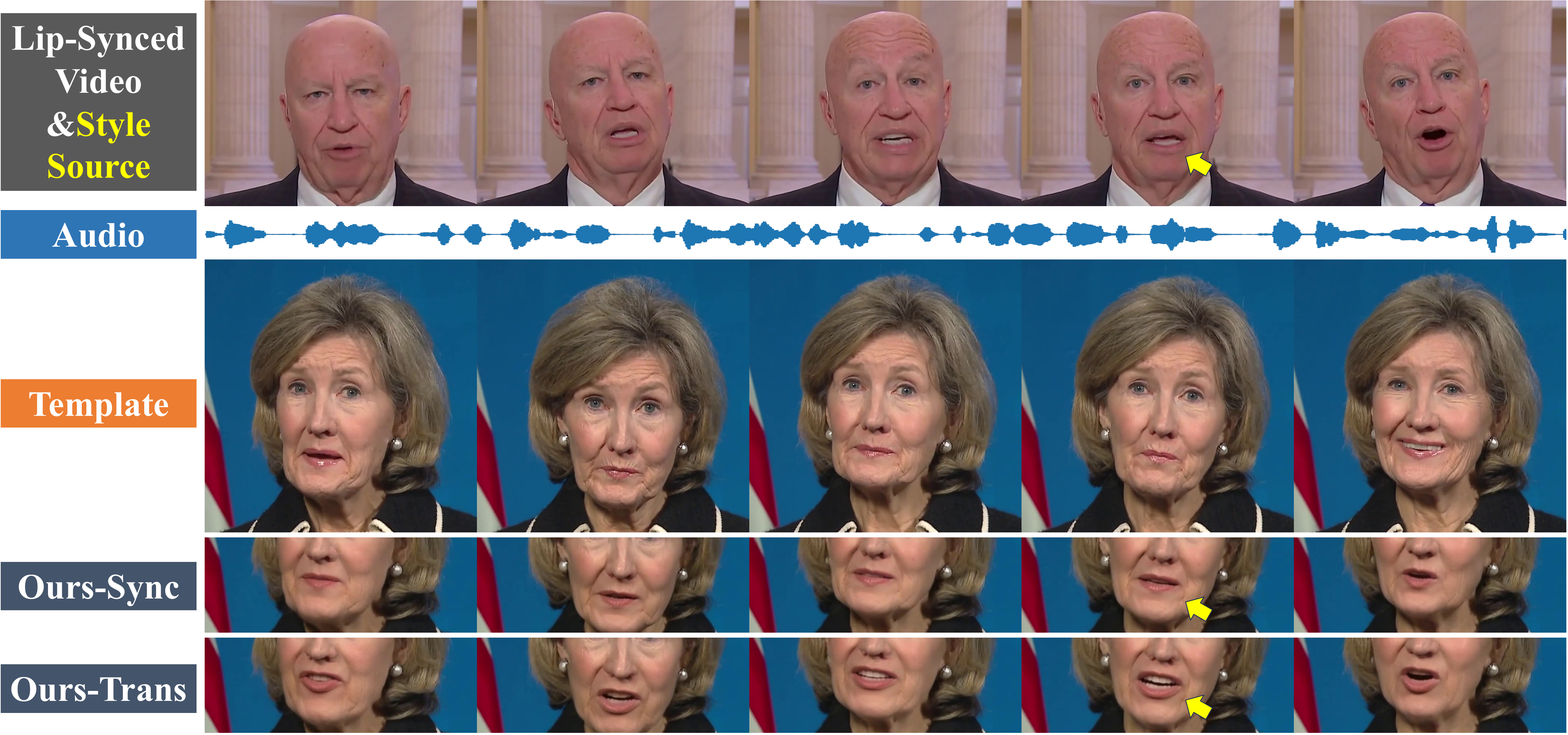}
\caption{\textbf{Style-Transferred Cross-Syncing.} ``Ours-Trans'' generates lip-synced results conditioned on the given audio and style source.}
\label{fig:show_trans}
\end{figure}
\begin{figure}[t]
\centering
\includegraphics[width=\linewidth]{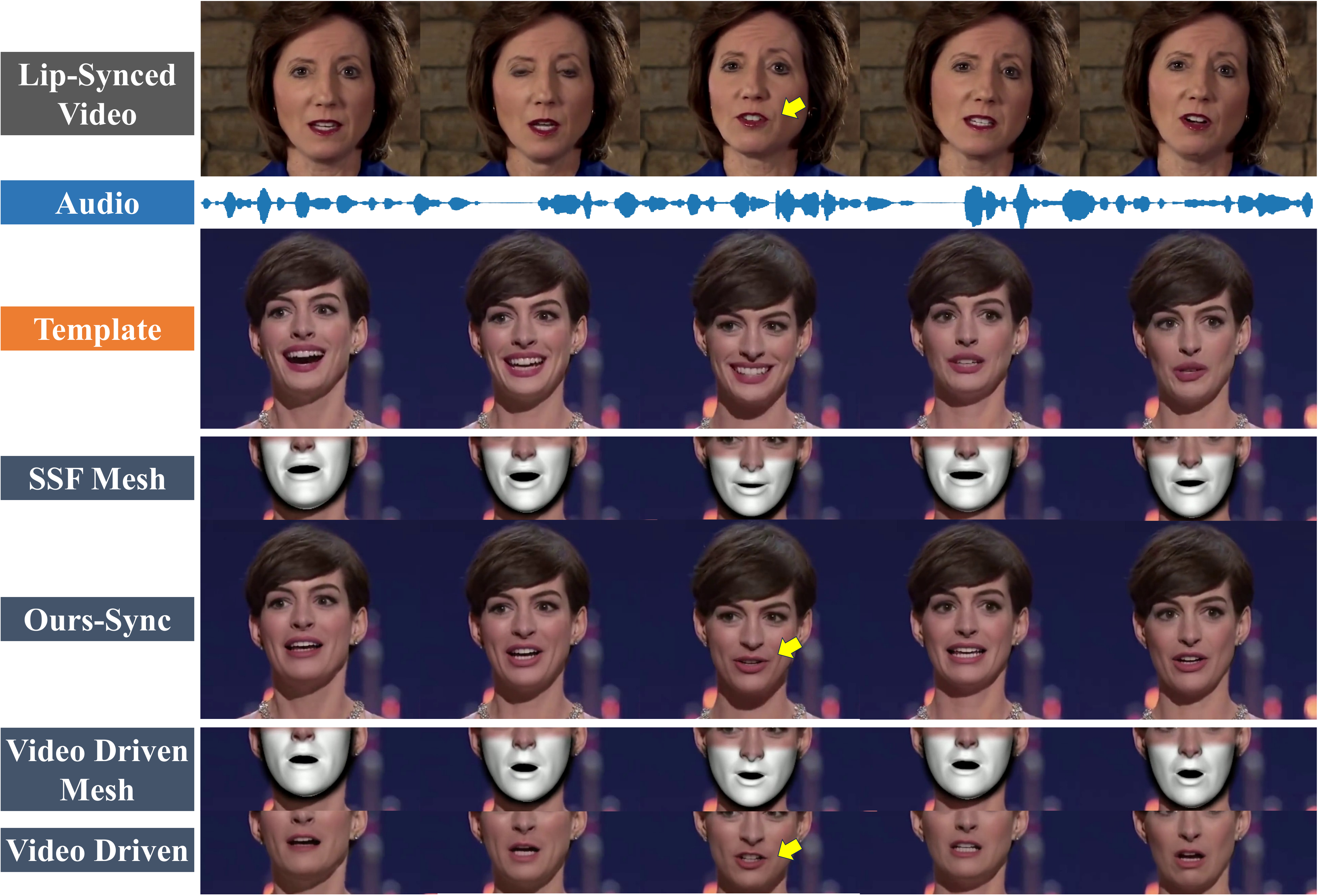}
\caption{\textbf{ Video Driven.} ``SSF Mesh'' denotes the results of the proposed Style-SyncFormer, which preserves person-specific attributes better.}
\label{fig:cmp_video_driven}
\end{figure}

\setlength{\tabcolsep}{5pt}
\begin{table}[t]
\centering
\caption{{Quantitative results on personalized test set.} For LMD and $\Delta{\text{Sync}}$ the lower the better, the higher the better for others.}
\resizebox{.7\linewidth}{!}{
\begin{tabular}{lcccc}
\toprule
\multirow{2}{*}{Method} &   \multicolumn{4}{c}{Dataset from~\cite{lu2021live}} \\ 
\cline{2-5}
\noalign{\smallskip}
                        & SSIM $\uparrow$ & PSNR $\uparrow$ & LMD $\downarrow$ & $\Delta{\text{Sync}}$ $\downarrow$  \\ 
\hline
\noalign{\smallskip}
LSP~\cite{lu2021live}  & - & -& -& 2.37 \\
RAD-NeRF~\cite{tang2022real}  & 0.63 & 29.65 & 8.16 & 2.23
 \\
StyleSync-P~\cite{guan2023stylesync} & 0.78 & 31.41 & 4.65 & \textbf{0.55}
 \\
GeneFace~\cite{ye2023geneface}  &  0.74 & 30.80 &  5.46 & 0.80
 \\
\hline
\noalign{\smallskip}
Ours-Sync-P & \textbf{0.80} & \textbf{31.44} & \textbf{4.35} &  \textbf{0.55} \\
\bottomrule
\end{tabular}
}
\label{tab:exp_cmp_self_sync_personalized}
\end{table}
\vspace{1cm}

\subsection{Comparison with Personalized Methods}
\label{sec:compare_personalized_method}

\noindent\textbf{Dataset.}
In order to compare with the SOTA personalized methods, we further fine-tune our model using 5 videos from the dataset introduced in~\cite{lu2021live}. Specifically, each video consists of 4500$\sim$6000 frames in fps 25. We keep the first 20 seconds of each video as the test set and leave the rest part for training.

\noindent\textbf{Comparison Method.}
We include four recent arts in the comparisons including LSP~\cite{lu2021live}, RAD-NeRF~\cite{tang2022real}, StyleSync~\cite{guan2023stylesync}, and GeneFace~\cite{ye2023geneface}.
StyleSync demonstrates a similar capability to fast adapt to given videos. For clarity in this section, we denote it as StyleSync-P. Similarly, our method is denoted as Ours-Sync-P to avoid confusion.

\noindent\textbf{Quantitative Result.}
The lip-sync evaluation following the same protocol as introduced in Sec.~\ref{sec:Quantitative Comparisons} is tabulated in Table.~\ref{tab:exp_cmp_self_sync_personalized}. 
Our method consistently outperforms recent SOTAs in both visual quality and sync quality. 
Please note that LSP generates lip-synced results with a random pose, rendering it impractical to access the ground truth images for calculating several metrics. 

\noindent\textbf{Qualitative Result.}
We present the comparisons using two randomly sampled driving audios in Fig.~\ref{fig:cmp_personalized_method}. 
In the illustration, it is evident that personalized methods yield high-quality results, preserving rich details around the mouth. Nonetheless, some methods struggle to accurately convey synchronized lip shapes, as we note in the figure with yellow arrows. 
Furthermore, we provide qualitative ablation and comparison on a less-than-30-second ``Trump video'' and a less-than-30-second ``Taylor video'' in our supplementary video. This showcases our method's ability to adapt to videos of less than 30 seconds.
Please refer to the video for a more intuitive comparison.

\begin{table}[t] 
\centering
\caption{{User study of personalized lip-sync.} The scores are ranged from 1 (worst) to 5 (best).}
\resizebox{.8\linewidth}{!}{
\begin{tabular}{cccccc}
\toprule
MOS on & {LSP} & {RAD-NeRF} & {StyleSync-P} & {GeneFace} & {{Ours-Sync-P}}
\\
\midrule
Lip-Sync Quality  &  3.79 & 2.75 & 4.21 & 3.33 & \textbf{4.58} \\
Generation Quality & 3.75 & 3.21 & 4.33 & 3.21 & \textbf{4.67} \\
Video Realness     & 3.25 & 2.42 & 3.96 & 2.58 & \textbf{4.62}  \\
\bottomrule
\end{tabular}
}
\label{table:sync_user_study_personalized}
\end{table}
\vspace{1cm}

\begin{table}[t]
\centering
\caption{{User study of face-swap.} The scores are ranged from 1 (worst) to 5 (best).}
\resizebox{.8\linewidth}{!}{
\begin{tabular}{cccccc}
\toprule
MOS on & {SimSwap} & {InfoSwap} & {StyleSwap} & {E4S} & {{Ours-Swap}} \\
\midrule
 Identity Similarity  & 2.83 & 2.46 & 3.38 & 2.92 & \textbf{3.83}\\
 Generation Quality & 2.88 & 1.71 & 3.17 & 1.92 & \textbf{3.92} \\
Video Realness     & 3.17 & 1.21 & 3.17 & 1.25 & \textbf{3.83} \\
\bottomrule
\end{tabular}
}
\label{table:swap_user_study}
\end{table}
\vspace{1cm}

\noindent\textbf{User Study.}
Similar to the user study introduced in Sec.~\ref{sec:Qualitative Comparisons}, we also conducted experiments regarding tasks of personalized lip-sync and face swapping.
We adopt the commonly used Mean Opinion Scores (MOS) rating protocol. All users give their ratings, ranging from 1 to 5, from three perspectives. 
For personalized lip-sync, the questions are set identically to the generalized setting as introduced in the paper.
For face swapping, participants are asked to evaluate: 
a) the identity similarity between the generated video and ID reference.
b) visual quality of the generated video, and 
c) whether this video looks real. 
The results averaged from 10 videos for each task are grouped into Table~\ref{table:sync_user_study_personalized} and Table~\ref{table:swap_user_study}, respectively.

From the tables, our method achieves the best performance over all three settings. 
Comparing the generalized and personalized lip-syncing, scores from three perspectives are generally improved with person-specific training. 
Nevertheless, the NeRF-based methods, RAD-NeRF and GeneFace obtain a relatively low ``video realness'' score. 
This highlights the limitation of the head-torso-separate rendering process, resulting in less vivid outcomes.

Regarding face-swap, our method shows surprisingly the highest ``identity similarity'', while StyleSwap performs slightly better in the quantitative evaluation (Table~\ref{table:cmp_swap_metrics}). We suspect that the improved generation quality and facial motion in our method lead experimenters to be more inclined to trust our results. 
Additionally, the scores for face-swap are lower than those in lip-sync, indicating that face-swap is a more challenging task and there is still room for improvement.

\subsection{Additional Experiments}
\label{sec:supp_additional_exp}

\subsubsection{Generalized 3D Dynamics Prediction.}
\begin{figure}[]
\setlength{\abovecaptionskip}{2pt}
\begin{minipage}{0.4\linewidth}
\centering
\includegraphics[width=\linewidth]{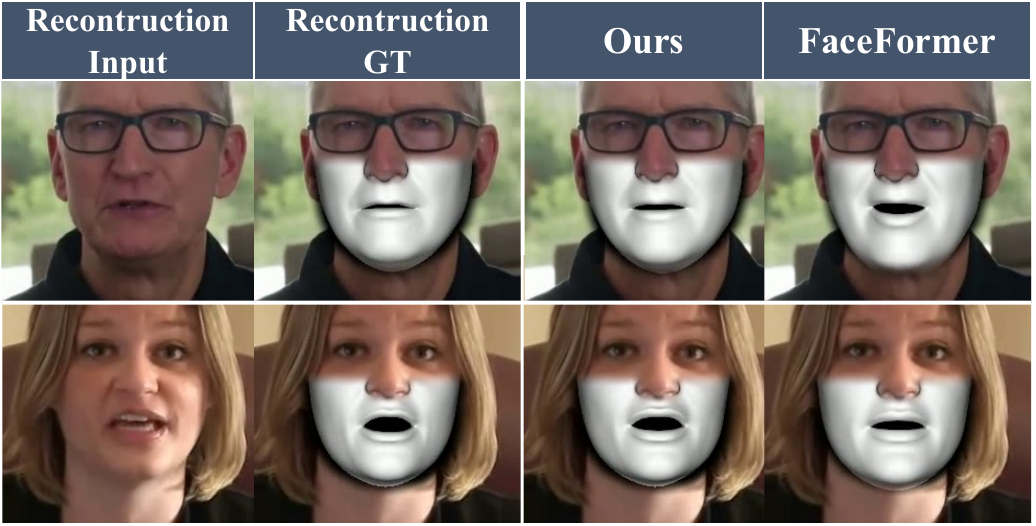}
\caption{\textbf{Qualatative comparison with FaceFormer.} 
}
\label{fig:cmp_faceformer}
\end{minipage}
\begin{minipage}{0.58\linewidth}
\centering
\captionof{table}{Quantitative comparison with FaceFormer on HDTF. The same second-stage generator is adopted for pixel space prediction. Clear gaps in LMD, $\Delta{\text{Sync}}$, and $\mathcal{D}_{\text{v}}$ show our superiority.
}
\resizebox{\linewidth}{!}{
\begin{tabular}{lccccc}
\toprule
{Method} & SSIM $\uparrow$ & PSNR $\uparrow$ & LMD $\downarrow$ & $\Delta{\text{Sync}}$ $\downarrow$ & $\mathcal{D}_{\text{v}}$ $\downarrow$  \\ 
\midrule
FaceFormer & 0.82 & 31.25 & 7.95 & 0.78 & 0.107 \\
Ours & \textbf{0.84} & \textbf{31.76} & \textbf{4.34} & \textbf{0.66} & \textbf{0.015} \\
\bottomrule
\end{tabular}
}
\label{tab:cmp_faceformer}
\end{minipage}
\end{figure}

Concerning the first stage, we compare the most related FaceFormer~\cite{fan2022faceformer} with simple necessary modifications. To adapt to unseen identities, the one-hot identity label of \cite{fan2022faceformer} is set to the same during training and inference. We re-train it on the same data as ours. The qualitative and quantitative results are shown in Fig.~\ref{fig:cmp_faceformer} and Table~\ref{tab:cmp_faceformer}. 
In the table, $\mathcal{D}_{\text{v}}$ denotes L2 error between the predicted vertices and GTs. Other metrics are introduced in Sec.~\ref{sec:Quantitative Comparisons}. 
Both visual results and the clear gaps in LMD, $\Delta{\text{Sync}}$, and $\mathcal{D}_{\text{v}}$ show our superiority in predicting more precise 3D dynamics with better person-specific characteristics. 
This is attributed to the speaking style learning and style-injected prediction process in our Style-SyncFormer.
While the close results in SSIM and PSNR are derived from the same second-stage generator.

\subsubsection{Facial Topology.}
\begin{figure}
\setlength{\abovecaptionskip}{2pt}
\begin{minipage}{0.45\linewidth}
\centering
\includegraphics[width=0.9\linewidth]{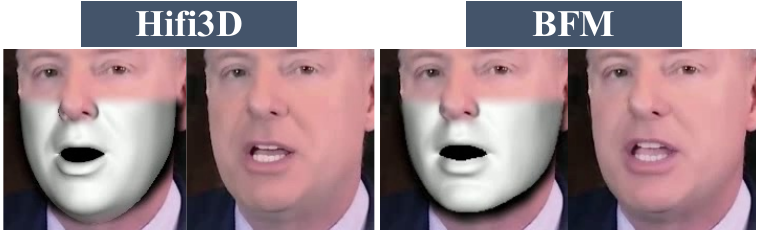}
\caption{\textbf{Qualatative comparison.} 
}
\label{fig:cmp_topo}
\end{minipage}
\begin{minipage}{0.54\linewidth}
\centering
\vspace{-0.15cm}
\captionof{table}{Quantitative comparison on HDTF.}
\resizebox{0.95\linewidth}{!}{
\begin{tabular}{lccccc}
\toprule
{Topology} & SSIM $\uparrow$ & PSNR $\uparrow$ & LMD $\downarrow$ & $\Delta{\text{Sync}}$ $\downarrow$ \\ 
\midrule
Hifi3D & {0.84} & {31.76} & {4.34} & {0.66} \\
BFM & 0.84 & 31.74 & 4.35 & 0.69 \\
\bottomrule
\end{tabular}
}
\label{tab:cmp_topo}
\end{minipage}
\end{figure}

As introduced, we reconstruct the 3D facial meshes with DeepFaceReconstruction~\cite{deng2019accurate} on the topology of Hifi3DFace~\cite{hifi3dface2021tencentailab} for the training of our first stage. Here we conduct ablations on the facial topology by comparing with BFM~\cite{bfm09}-based generation. Results are demonstrated in Fig.~\ref{fig:cmp_topo} and Table~\ref{tab:cmp_topo}. It shows that using different topologies in our Resyncer framework makes little difference.

\setlength{\columnsep}{5pt}%
\setlength{\intextsep}{4pt}%

\subsection{Supplementary Video}
The supplementary video is comprised of 4 sections including 1) a short introduction, 2) a demonstration of our performance and comparison with SOTA methods, 3) ablation studies, and 4) more intriguing results. 
\begin{itemize}
\item[1)] We demonstrate the capabilities of our method to generate lip-synced, speaking-style-transferred, and identity-swapped videos.
\item[2)] The comparisons are grouped into 3 parts including a) generalized lip-sync, b) personalized lip-sync, and c) face-swap.
\item[3)] We further provide video ablation results for a more intuitive demonstration.
\item[4)] More intriguing results including generated results based on template videos from the Internet. We further provide a comparison with a commercial tool (HeyGen) for the task of video translation.
\end{itemize}
Please note only videos with ``personalized''  listed (e.g. 2:00 - 3:05 in the video) and comparisons with HeyGen are finetuned.

\subsection{Limitations}
\begin{figure}
\centering
\includegraphics[width=.7\linewidth]{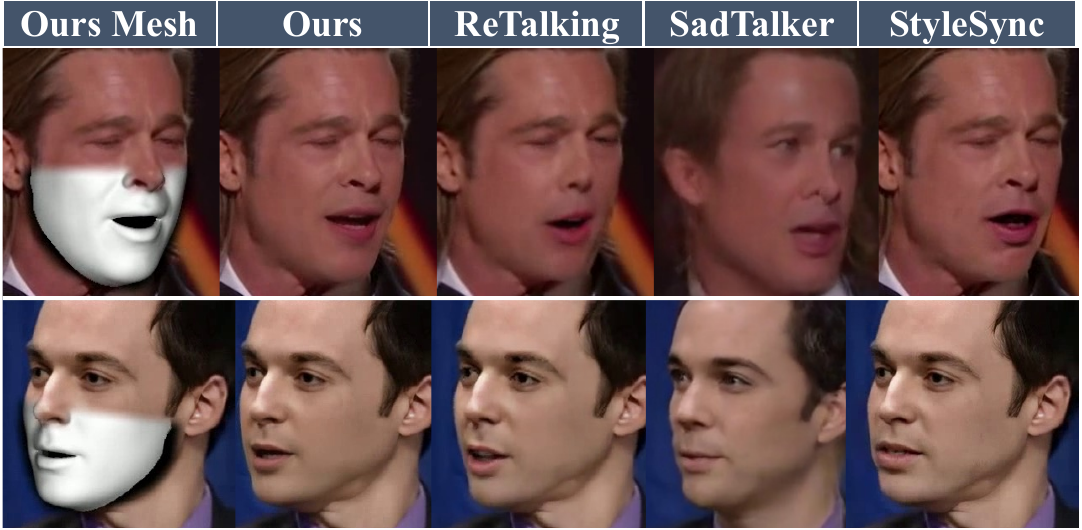}
\caption{\textbf{Comparison on large pose.} }
\label{fig:cmp_large_pose}
\end{figure}

Our framework might not work well on extreme poses (over 80 degrees). As our framework relies on the poses of the 3D facial meshes to formulate the input image, the error of face reconstruction at extreme poses might harm our final results. Here we further present a $\sim$50-degree comparison in Fig.~\ref{fig:cmp_large_pose}, showing superior performance of ours compared with SOTA methods.

